\documentclass[conference]{IEEEtran}
\IEEEoverridecommandlockouts
\usepackage{cite}
\usepackage{amsmath,amssymb,amsfonts}
\usepackage{algorithmic}
\usepackage{graphicx}
\usepackage{textcomp}
\usepackage{easy-todo}
\usepackage{comment}
\usepackage{xcolor}
\def\BibTeX{{\rm B\kern-.05em{\sc i\kern-.025em b}\kern-.08em
    T\kern-.1667em\lower.7ex\hbox{E}\kern-.125emX}}
\begin{document}

\title{Human eye inspired log-polar pre-processing for neural networks}

\author{
\IEEEauthorblockN{1\textsuperscript{st} Leendert A Remmelzwaal}
\IEEEauthorblockA{\textit{Department of Electrical Engineering}
\textit{University of Cape Town,}
Cape Town, South Africa\\
leenremm@gmail.com} \\
\IEEEauthorblockN{2\textsuperscript{nd} Amit Kumar Mishra}
\IEEEauthorblockA{\textit{Department of Electrical Engineering}
\textit{University of Cape Town,}
Cape Town, South Africa\\
akmishra@ieee.org} \\
\IEEEauthorblockN{3\textsuperscript{rd} George F R Ellis}
\IEEEauthorblockA{\textit{Department of Mathematics and Applied Mathematics}
\textit{University of Cape Town,}
Cape Town, South Africa\\
george.ellis@uct.ac.za}
}

\maketitle

\begin{abstract}
In this paper we draw inspiration from the human visual system, and present a bio-inspired pre-processing stage for neural networks. We implement this by applying a log-polar transformation as a pre-processing step, and to demonstrate, we have used a naive convolutional neural network (CNN). We demonstrate that a bio-inspired pre-processing stage can achieve rotation and scale robustness in CNNs. A key point in this paper is that the CNN does not need to be trained to identify rotation or scaling permutations; rather it is the log-polar pre-processing step that converts the image into a format that allows the CNN to handle rotation and scaling permutations. In addition we demonstrate how adding a log-polar transformation as a pre-processing step can reduce the image size to ~20\% of the Euclidean image size, without significantly compromising classification accuracy of the CNN. The pre-processing stage presented in this paper is modelled after the retina and therefore is only tested against an image dataset.\footnote{Note: This paper has been submitted for SAUPEC/RobMech/PRASA 2020}\\
\end{abstract}


\section*{Introduction}

We are all familiar with Euclidean model images whether we are aware of it or not: all of our digital photos and videos are in this format. Euclidean images have a width ($w$) and a height ($h$), and the total number of pixels (data points) can be calculated as ($w \times h$). Modern optical cameras typically capture images in Euclidean format, and visual processing algorithms typically process Euclidean images (e.g. OpenCV template matching). The Euclidean format for image processing has become the most prevalent model in image processing today.

\begin{figure}[htbp]
\centerline{\includegraphics[scale=0.45]{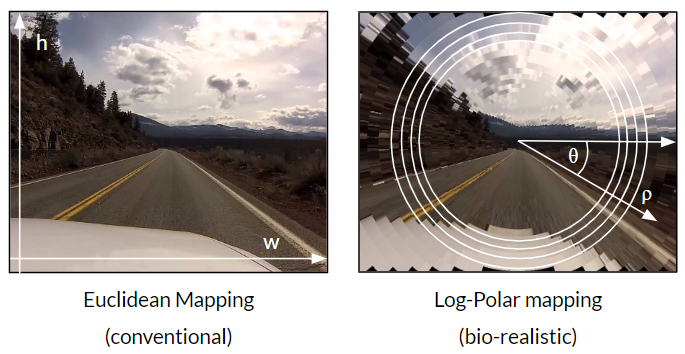}}
\caption{Euclidean and Log-Polar transformation of the same image.}
\label{Fig02}
\end{figure}

However the human brain does not processes the visual senses as a Euclidean image, but rather as a log-polar model \cite{wilson1992log}: a more bio-realisic model (see Fig~\ref{Fig02}) inspired by the structure of the fovea and retina of the human eye \cite{marr1982vision} \cite{hubel1995eye}. The human eye has many benefits over traditional optical cameras, including the fact that the human eye can handle object classification even when an object has been rotated or resized. In this paper we explore how bio-inspired architectures can enhance CNN performance, specifically relating to how CNNs classify objects that have been rotated or resized.

\subsection*{Log-polar transformations}

The log-polar model can be created from a Euclidean model by remapping points from the 2D Cartesian coordinate system (x,y) to the 2D log-polar coordinate system ($\rho$, $\theta$) as follows:

\begin{equation}
\rho = \log \left( (y - y_c)^2 + (x - x_c)^2 \right) \\
\end{equation}

\begin{equation}
\theta = \arctan \left( \frac{y - y_c}{x - x_c} \right) \, \text{where} \, x > 0
\end{equation}

where the center of the image is $(x_c, y_c)$, $\theta$ is the angle or rotation, and $\rho$ is the logarithm of the distance from the center $(x_c, y_c)$ to the data point $(x, y)$.

It remains a challenge to calculate rotation and scale efficiently from Euclidean images. Typical implementations of detecting rotation consist of running a Euclidean template match against the target image, and rotating the image multiple times to find which rotation results in the best match. Similarly for detecting scale, a common approach is to match templates of various sizes against the target image, and selecting the template size with the highest match correlation score.

\begin{figure}[htbp]
\centerline{\includegraphics[scale=0.4]{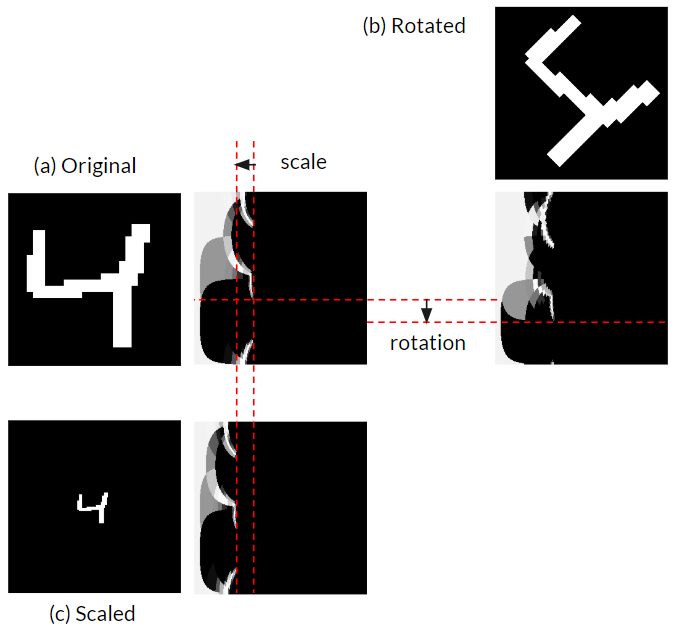}}
\caption{Rotation and Scaling transformations to a Euclidean image can be read as horizontal and vertical shift respectively, after a log-polar transformation. The log-polar transformation translates rotation and scale in Euclidean images into vertical and horizontal translations (respectively) in the log-polar model.}
\label{Fig03}
\end{figure}

The log polar transformation of a Euclidean models has the significant advantage of converting rotation and scale transformations into a format that can be easily processed with Euclidean algorithms. As shown in Fig~\ref{Fig03}, the rotation of a Euclidean image can be read as a horizontal shift in the image after the log-polar transformation, and the scale of the image can be similarly read as a vertical shift.

\subsection*{Related Work}

Log-polar transformations have been used to estimate skewness of text in an image \cite{brodic2013log}, to detect road signs in an image \cite{ellahyani2017mean}. Amorim et al. \cite{amorim2018analysing} demonstrated that log-polar pre-processing can assist with rotation invariance in CIFAR10 images, but did not demonstrate scale invariance, data processing efficiency, nor were they able to demonstrate rotation invariance with the character-based datasets (e.g. MNIST dataset). Algorithms such as the Fourier-Mellin algorithm \cite{sheng1986circular} use log-polar representation to approximate the rotational difference between two images. To explore the value of log-polar models, foveated image sensors have been designed \cite{wodnicki1995foveated}, and log-polar transformation algorithms are also available (e.g. OpenCV). Log-polar transformation have also been used neural network pre-processing \cite{henriques2017warped} to estimate the scale and rotation of an image.

In this paper we build on the existing implementations of log-polar transformations, and explore how the log-polar transformation can improve classification in convolutional neural networks (CNNs) by adding (1) rotation-invariance, (2) scale-invariance, and (3) image compression.

\subsection*{Aim of this paper}

We aim to demonstrate that a log-polar pre-filter and post-filter will improve the CNN's ability to correctly classify images with rotation and scaling transformations. In addition, we aim to demonstrate that applying a log-polar transformation as a pre-processing step to a classification CNN can significantly reduce the image size required by the CNN without significantly compromising the classification accuracy of the model.

\subsection*{Structure of this paper}

In section 1 we describe the architecture of the log-polar model and neural network. In section 2 we describe the simulations we ran and the observations made, and then draw conclusions in Section 3.

\section{Model architecture}

\subsection*{Neural Network}

\begin{figure}[htbp]
\centerline{\includegraphics[scale=0.4]{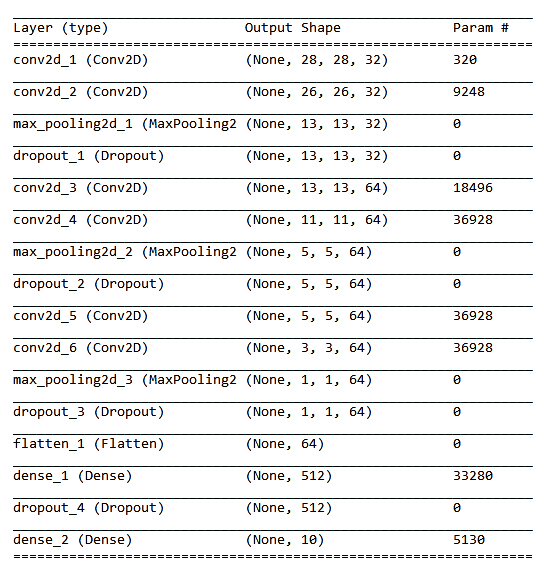}}
\caption{We used a CNN architecture with 784 input nodes (28x28 pixels) and 10 output nodes (10 digit classes), with a selection of convolutional, dropout and pooling layers.}
\label{Fig05}
\end{figure}

We chose a translation-invariant CNN to classify the images because we after the log-polar transformation we still require the convolution operation to extract the high-level features: a CNN is equivariant to image translation\cite{henriques2017warped}. We chose a CNN architecture with 784 input nodes (28x28 pixels) and 10 output nodes (10 digit classes), with a selection of convolutional, dropout and pooling layers, as shown in Fig~\ref{Fig05}.

\subsection*{Experimental Design}

As shown in Fig~\ref{Fig04}, four experiments were designed: (a) the first one as a control using the standard MNIST dataset and a CNN as a classifier, (b) the second one with using the standard MNIST dataset and a log-polar pre-filter, (c) the third one had rotation and scaling applied to the standard MNIST dataset, but using the same trained CNN as in (a), and the fourth one with a log-polar pre-filter on the same CNN used in (b).

\subsection*{Dataset}

We use the MNIST database of handwritten digits \cite{yann1998mnist}, consisting of 70,000 labelled, 28x28 grayscale images of the 10 handwritten digits (0-9). The training set consists of 60,000 images while the testing set consists of 10,000 images. We chose the MNIST dataset because the log-polar transformation works best with (a) square images, where (b) the object has been centered in the frame, which is the case with the MNIST dataset.

\begin{figure}[htbp]
\centerline{\includegraphics[scale=0.5]{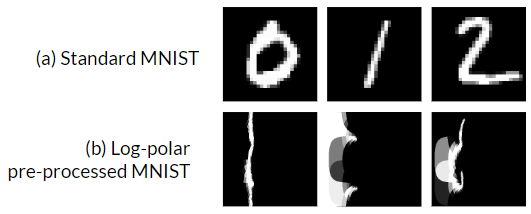}}
\caption{Two variations of the MNIST dataset were used: one using the standard Euclidean format, and the other with a log-polar transformation applied to each image.}
\label{Fig07}
\end{figure}

Two variations of the MNIST dataset were used: one using the standard Euclidean format, and the other with a log-polar transformation applied to each image (as shown in Fig~\ref{Fig07}).

\begin{figure}[htbp]
\centerline{\includegraphics[scale=0.25]{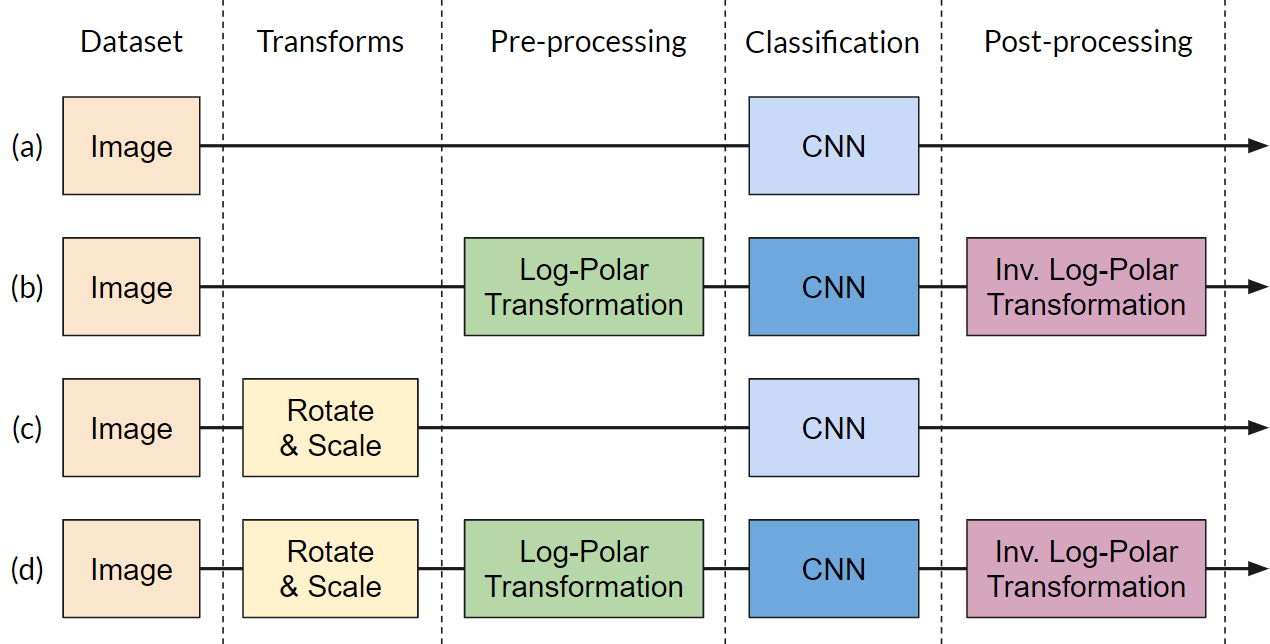}}
\caption{We designed four experiments: (a) the first one as a control using the standard MNIST dataset (orange) and a CNN (light blue) as a classifier, (b) the second one with using the standard MNIST dataset and a log-polar pre-filter (green), (c) the third one had rotation and scaling (yellow) applied to the standard MNIST dataset, but using the same trained CNN as in (a), and the fourth one with a log-polar pre-filter (green) on the same CNN used in (b).}
\label{Fig04}
\end{figure}

To control for rotation and scaling variations in our experiment, we controlled the rotation and scaling of each image in the MNIST dataset; this is shown in Fig~\ref{Fig04} as the ``Transforms" phase. 

\section*{Results}

\subsection*{Experiment A \& B: Control Experiments}

As shown in in Fig~\ref{Fig04}, the experiments A and B served a control, before rotation and scaling transformations were applied to the dataset. The purpose of these experiments were to see how the CNN performed with classifying the standard labelled MNIST dataset, as well as log-polar transformation of the labelled MNIST data.

We trained two separate CNNs on variations of the MNIST datasets: (1) the standard MNIST dataset for use in experiment A and C, and (2) the MNIST dataset with log-polar pre-processing applied for use in experiment B and D. 

\begin{figure}[htbp]
\centerline{\includegraphics[scale=0.5]{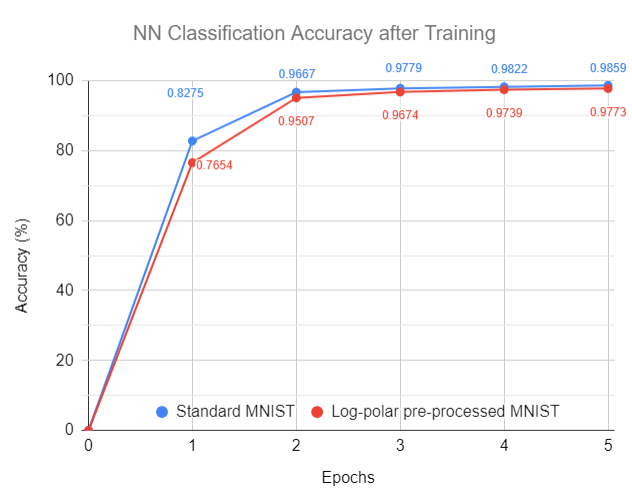}}
\caption{Neural network training accuracy results on two variations of the MNIST datasets: (1) the standard MNIST dataset for use in experiment A and B, and (2) the MNIST dataset with log-polar pre-processing applied for use in experiment C. After training with only 5 epochs, both CNN models achieved a classification accuracy of above 97\%.}
\label{Fig06}
\end{figure}

After training with only 5 epochs, both CNN models achieved a baseline classification accuracy of above 97\% (see Fig~\ref{Fig06}). The CNN trained on the standard MNIST dataset achieved a slightly higher classification accuracy of 98.59\%, while the CNN trained on the log-polar transformed MNIST dataset achieved classification accuracy of 97.73\% after only 5 epochs of training.

\subsection*{Experiment C \& D: Rotation and Scale}

Having established an accuracy baseline in experiment A and B, we then introduced rotational and scaling transformations to the dataset. For both experiments C and D, neither of the CNNs were re-trained; instead the CNN's retained their training from experiments A and B were used, respectively. The reason for doing this was to see how well the CNN models classified unseen images with rotational and scaling transformations applied, and whether the log-polar transformation assisted the CNN with image classification.

\begin{figure}[htbp]
\centerline{\includegraphics[scale=0.25]{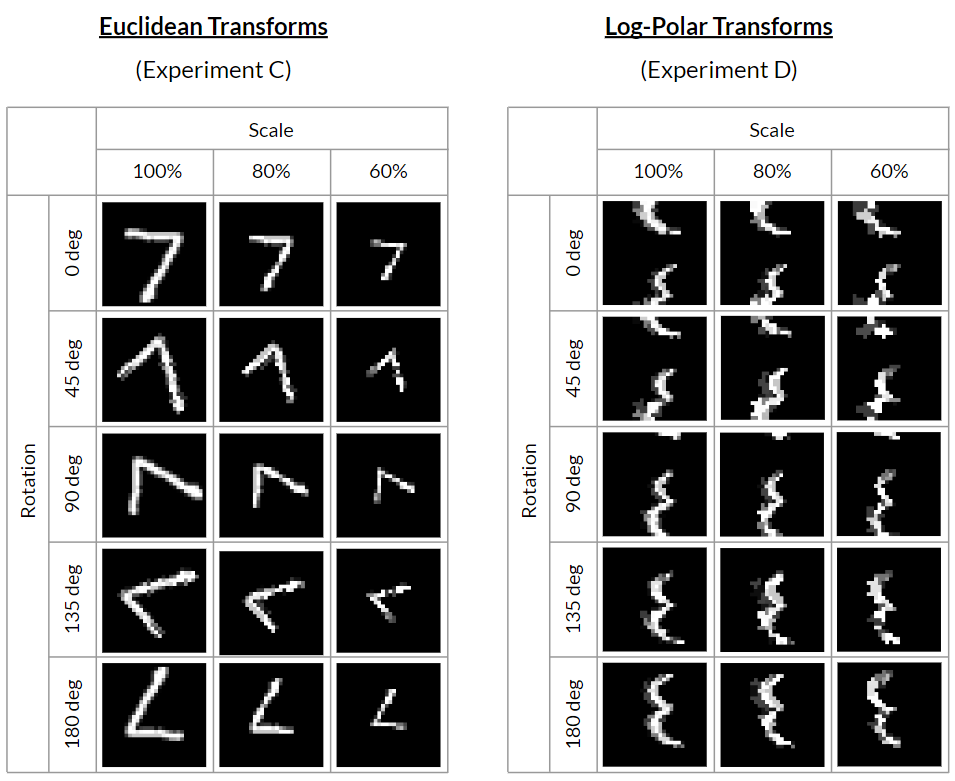}}
\caption{Example of rotation and scale transformations for a single MNIST image. The Euclidean transforms were used in experiments A and C, and the log-polar transforms were used in experiments B and D.}
\label{Fig10}
\end{figure}

For experiment C and D, we varied the rotation of each image from 0 degrees to 360 degrees, and we varied the scale from 100\% - 40\%. Examples of these transformations for both datasets are shown in Fig~\ref{Fig10}. 

\begin{figure}[htbp]
\centerline{\includegraphics[scale=0.7]{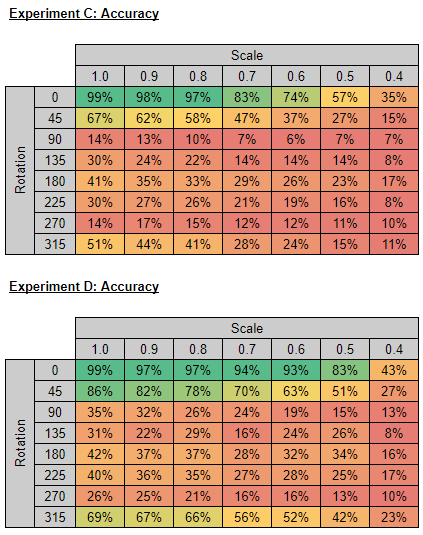}}
\caption{The accuracy of the CNN for permutations of rotation and scale, for both the Euclidean images as well as the log-polar images.}
\label{Fig08}
\end{figure}

The accuracy of the CNN was recorded for permutations of rotation and scale, for both the Euclidean images as well as the log-polar images (see Fig~\ref{Fig08}). Firstly we observed that the CNN's accuracy when processing Euclidean images declined sharply when a rotation or scale transformation was applied to the images. The CNN managed to maintain an accuracy greater than 90\% only when the scale was varied between 80\% and 100\%. By comparison, we observed that the CNN's performed significantly better when processing log-polar images: the CNN managed to maintain an accuracy greater than 90\% only when the scale was varied between 60\% and 100\%. In addition, we observed that the CNN had great accuracy in classifying images that had been rotated.

\begin{figure}[htbp]
\centerline{\includegraphics[scale=0.7]{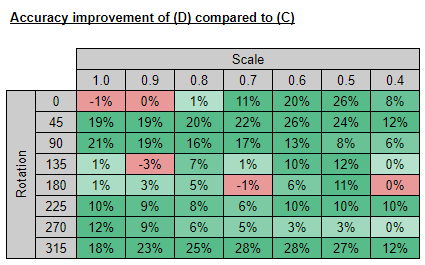}}
\caption{The performance improvement of the CNN that processed log-polar images, compared to the CNN that processed Euclidean images (see Fig~\ref{Fig08}).}
\label{Fig09}
\end{figure}

To highlight the performance improvement, we mapped the performance improvement of the CNN that processed log-polar images, compared to the CNN that processed Euclidean images (see Fig~\ref{Fig09}). We observed that the rotational accuracy was significantly higher for rotations in the range -90 degrees to 90 degrees. We also observed that the accuracy remained high for variations of scale from 50\% to 100\% also for rotations in the range -90 degrees to 90 degrees.

\subsection*{Log-polar Image Compression}

\begin{figure}[htbp]
\centerline{\includegraphics[scale=0.42]{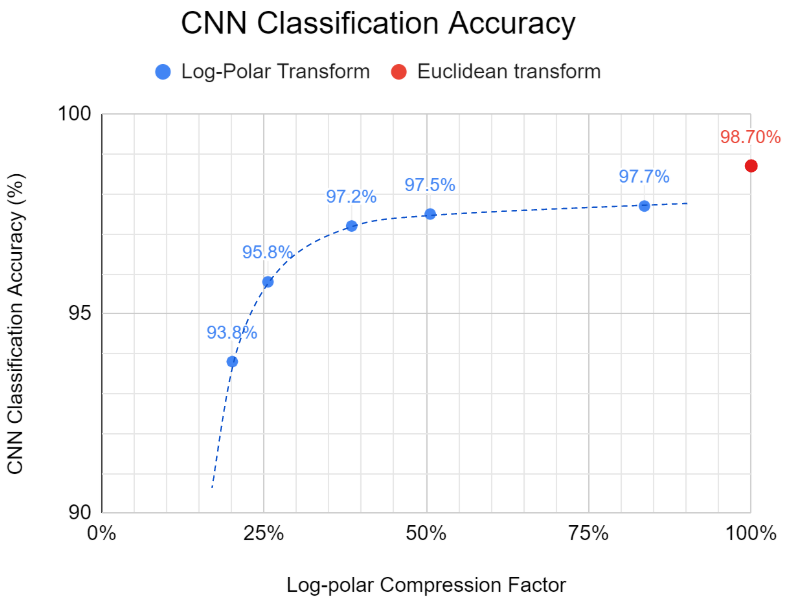}}
\caption{The CNN classification accuracy for varying log-polar compression factors: the log-polar compression factor was varied, and the CNN was re-training every time with 5 training epochs. We observed that a log-polar compression as low as 20.1\% still resulted in an overall classification accuracy of 93.8\%.}
\label{Fig11}
\end{figure}

It was observed that the log-polar transformation also applied a compression to the images in the datasets, reducing their overall size. The log-polar compression factor was varied, and the CNN was re-training every time with 5 training epochs. We observed that a log-polar compression as low as 20.1\% still resulted in an overall classification accuracy of 93.8\%, which is still very high compared to the benchmark CNN accuracy of 98.7\%. The results from the simulations are shown in Fig~\ref{Fig11}.

\section*{Conclusion}

In this paper we show that the log-polar transformation can have a positive impact on the performance and accuracy of a CNN. It is not altogether surprising to see this, given that the log-polar model is biologically similar to the way the retina in the human eye receives information, and how the cortex efficiently processes this information.

\noindent{We make the following conclusions:}

\begin{enumerate}
\item We observed that adding the log-polar transformation as a pre-processing step encodes the rotation and scale transforms into an image format that can be processed easily by a CNN.
\item We observed that log-polar pre-processing allows for a CNN to handle rotation and scaling transformations efficiently, even if the CNN has not been previously trained to process rotation or scaling permutations of images.
\item We also observed that applying a log-polar transformation can compress the image size, while maintaining a high CNN classification accuracy.
\end{enumerate}
We have demonstrated that applying a log-polar transformation as a pre-processing step to a classification CNN can reduce the image size required by the CNN, and allow for improved performance in handling rotation and scaling permutations of images.

\subsection*{Future Work}

The model presented in this paper is modelled after the retina and therefore is only demonstrated using an image dataset (MNIST). However, the same techniques can also be applied to other datasets.

In this paper we observed a positive rotation-invariant response for rotations between -90 degrees, and +90 degrees. If we were to train the CNN on the same MNIST dataset, but also on a set of images rotated 180 degrees, we hypothesize that we may see a rotation-invariant response for the entire 360 degree rotation: 0 degrees +- 90 degrees, combined with 180 degrees +- 90 degrees. This hypothesis has yet to be tested and proven.

\subsection*{Supporting Material}

The source code as well as records of the tests conducted in this paper are publicly available online \cite{remmelzwaal2019logpolar}. For additional information, please contact the corresponding author.

\subsection*{Acknowledgements}

\noindent{This research did not receive any specific grant from funding agencies in the public, commercial, or not-for-profit sectors.}

\end{document}